\documentclass{article}

\usepackage{macros_hristo}

\usepackage[dblblindworkshop,final]{neurips_2025}

\workshoptitle{MATH-AI}

\usepackage[utf8]{inputenc} 
\usepackage[T1]{fontenc}    
\usepackage{hyperref}       
\usepackage{url}            
\usepackage{booktabs}       
\usepackage{amsfonts}       
\usepackage{nicefrac}       
\usepackage{microtype}      
\usepackage{xcolor}         

\title{Exact Learning of Arithmetic with Differentiable Agents}

\author{Hristo Papazov \\
TML Lab, EPFL \\
\texttt{hristo.papazov@epfl.ch}
\And Francesco D'Angelo \\
TML Lab, EPFL \\
\texttt{francesco.dangelo@epfl.ch}
\And Nicolas Flammarion \\
TML Lab, EPFL \\
\texttt{nicolas.flammarion@epfl.ch}
}

\begin{document}

\maketitle

\vspace{-5mm}
\begin{abstract}
We explore the possibility of exact algorithmic learning with gradient-based methods and introduce a differentiable framework capable of strong length generalization on arithmetic tasks. Our approach centers on Differentiable Finite-State Transducers (DFSTs), a Turing-complete model family that avoids the pitfalls of prior architectures by enabling constant-precision, constant-time generation, and end-to-end log-parallel differentiable training. Leveraging policy-trajectory observations from expert agents, we train DFSTs to perform binary and decimal addition and multiplication. Remarkably, models trained on tiny datasets generalize without error to inputs thousands of times longer than the training examples. These results show that training differentiable agents on structured intermediate supervision could pave the way towards exact gradient-based learning of algorithmic skills. Code available at \href{https://github.com/dngfra/differentiable-exact-algorithmic-learner.git}{https://github.com/dngfra/differentiable-exact-algorithmic-learner.git}.
\end{abstract}

\vspace{-3.5mm}
\section{Introduction}\label{sec:intro}
\vspace{-3.0mm}
The dream of AGI envisions systems capable of exact learning of algorithmic skills -- such as searching, sorting, or arithmetic -- from a reasonable dataset of examples \citep{solomonoff2006machine, chollet2019measure, gyorgy2025beyond}. Here, \textit{exact learning} refers to the criterion proposed by \citet{angluin1988queries} which requires correct application of a ground-truth rule to every possible input. More formally, given a finite problem alphabet $\Sigma$ and an unknown model $M^\star$ computing some function $f^\star : \D \to \SS $ over an input domain $\D \subseteq \SS$, the exact learning paradigm requires a learner $\L$ to identify $f^\star$ with probability 1 after observing some part of $M^\star$’s computation on finitely many inputs $z_1, \dots, z_N \in \D$.

In this paper, we entertain the idea of gradient-based exact learning and provide as evidence a simple differentiable agent capable of unprecedented state-of-the-art length generalization on arithmetic tasks. In pursuing the ambitious goal of differentiable exact learning, a couple of considerations arise.

\textbf{Training Data.} First, we cannot expect our differentiable learner $\L$ to identify every computable function $f^\star$ from input-output examples $(z, f^\star(z))$ alone. Indeed, \citet{gold1967language} showed that no algorithm can exactly learn the class of recursive functions in the limit from input–output examples (Theorem I.5), highlighting the inadequacy of such training data. Consequently, we adopt the formal framework of \citet{papazov2025learning} for learning algorithms in the limit from observations of intermediate steps. As we will soon discuss, the addition of observable intermediate steps, termed \textit{policy-trajectory observations} (PTOs), allows us to reduce the intractable problem of learning recursive functions to the more manageable one of learning finite-state transducers (FSTs) \citep[see Theorem 16]{papazov2025learning}.

\textbf{Model Class.} Second, since for any computable $f^\star$, we want gradient-based search to find some differentiable model capable of executing $f^\star$, we need to ensure the computationally universality (or Turing-completeness) of our model class. To the best of our knowledge, all previously proposed universal model families commit at least one of the following two unforgivable sins for optimization:
\begin{itemize}[label = $\bullet$, topsep=5pt, left=5pt, itemsep=3pt, labelsep=5pt]
    \item \textbf{Exploding Computation Time:} RNNs and Neural GPUs with unbounded precision of internal arithmetic \citep{siegelmann1992computational, perez2018on}; Transformers with growing context windows and unbounded/scaling internal arithmetic \citep{perez2018on, merrill2023expresssive, bhattamishra2020computational}; recurrent models with growing differentiably referenceable memories \citep{grefenstette2015learning};
    \item \textbf{Uninformative Gradients:} Architectures with non-differentiable adaptive-computation-time mechanisms \citep{graves2016adaptive, dehghani2018universal}; models interacting with an external environment without differentiable (or even continuous) feedback \citep{joulin2015inferring, chung2021turing, schuurmans2024autoregressive}.
\end{itemize}
These caveats hurt gradient-based optimization. The first not only causes an inference-time explosion but also increases the computational distance between emitting an output and receiving a gradient signal. The second produces gradients with dubious optimization information in a discontinuous loss landscape. Thus, the cited works achieve Turing-completeness at the cost of rendering gradient-based training impossible. In contrast, by allowing our Differentiable Finite-State Transducer (DFST) family free motion over an external environment, we addresses all three limitations and present a framework with constant precision, constant-time output generation, computational universality, and the capacity for end-to-end differentiable training.

Note that we omit any optimization critique of other neurosymbolic architectures such as Neural Turing Machines \citep{graves2014neural}, Neural GPUs \citep{kaiser2015neural}, Neural Random-Access Machines \citep{kurach2015neural}, Neural Programmer-Interpreters \citep{reed2015neural}, Pointer Networks \citep{vinyals2015pointer}, Hierarchical Attentive Memories \citep{andrychowicz2016learning}, finite-precision recurrent networks \citep{weiss2018practical, merrill2019sequential}, finite-precision Transformers \citep{perez2018on, hahn2020theoretical, merrill2023parallelism}, SSMs \citep{merrill2024illusion}, and Hierarchical Reasoning Models \citep{wang2025hierarchical} since these model families lack computational universality and therefore cannot, in principle, exactly learn arbitrary algorithms. Moreover, most of the listed architectures can only express linear memory algorithms.

\textbf{Main Contributions.} Our paper introduces a simple, differentiable, Turing-complete, and parallelizable setup consisting of a DFST agent which interacts with an external environment through action tokens. We train our DFST model on observed computational traces (PTOs) from expert agents performing four different arithmetic tasks: binary addition and multiplication (\texttt{add2}, \texttt{mult2}) and decimal addition and multiplication (\texttt{add10}, \texttt{mult10}) of two numbers. For \texttt{add2} and \texttt{add10}, we train on tiny datasets consisting of 20 and 225 examples, respectively, featuring summands with at most 3 digits. Training on these datasets led to both robust and probabilistic length generalization (RLG and PLG)\footnote{We formally define RLG and PLG for arithmetic tasks in \Cref{sec:exp}.} of 3850 and 2450 digits, respectively. Testing for correctness beyond these numbers of digits leads to \texttt{OutOfMemory} errors on our A100-SXM4-80GB GPU. Similarly, for \texttt{mult2} and \texttt{mult10}, we train on datasets consisting of 750 and 10000 samples, respectively, featuring multiplicands with up to 5 digits. Training on these datasets led to RLG and PLG of 600 and 180 digits, respectively, and we could not test on larger numbers due to insufficient memory. In particular, as far as we tested, we could not find a single error in the computation of our best trained models for each arithmetic task. Moreover, we achieved this state-of-the-art length generalization with a well-motivated scratchpad data in the form of PTOs and a straightforward training procedure.

\begin{table}[h!]
\centering
\begin{threeparttable}
\caption{Neural GPU vs. DFST on Arithmetic Tasks}
\label{tab:comparison}
\begin{tabular}{clcccc}
\toprule
Framework & Metric & \texttt{add2} & \texttt{add10} & \texttt{mult2} & \texttt{mult10} \\
\midrule
\multirow{2}{*}{Neural GPU} & \# Samples & $\sim200$k & N/A & $\sim200$k & N/A \\
& \# Model Parameters & 10368 & N/A & 10368 & N/A \\
\multirow{3}{*}{w/ Input-Output Data} & Max Train Number Length & 20 & N/A & 20 & N/A \\
& Robust LG & $\leq$ 20 & N/A & $\leq$ 8 & N/A \\
& Probabilistic LG & $\geq$ \textbf{2000} & N/A & $\geq$ \textbf{2000} & N/A \\
\midrule
\multirow{2}{*}{DFST (ours)} & \# Samples & 20 & 225 & 750 & 10000 \\
& \# Model Parameters & 1020 & 8900 & 5280 & 162108 \\
\multirow{3}{*}{w/ PTO Data} & Max Train Number Length & 3 & 3 & 5 & 5 \\
& Robust LG & $\geq$ \textbf{3850} & $\geq$ \textbf{2450} & $\geq$ \textbf{600} & $\geq$ \textbf{180} \\
& Probabilistic LG & $\geq$ \textbf{3850} & $\geq$ \textbf{2450} & $\geq$ \textbf{600} & $\geq$ \textbf{180} \\
\bottomrule
\end{tabular}
\begin{tablenotes}
\centering
\small
\item Bold \textbf{numbers} indicate inability to test on longer inputs due to GPU memory constraints.
\end{tablenotes}
\end{threeparttable}
\end{table}

\textbf{Related Work.}
Since exact learning represents the limit of length generalization, we review in passing some experimental results on length generalization in sequence-to-sequence models.
Recent empirical studies \citep{joulin2015inferring, deletang2022neural, liu2022transformers, shen2023positional, kazemnejad2023impact, jelassi2023length} demonstrated that Transformers, modified RNNs, and LSTMs length-generalize poorly on a variety of simple algorithmic tasks when trained on input-output data. With additional data-formatting strategies -- such as task-specific index hints, optimized positional encodings, input reversing, position coupling, and scratchpad guidance -- multiple works on arithmetic generalization in Transformers \citep{anil2022exploring, jelassi2023length, zhou2023algorithms, lee2023teaching, zhou2024transformers, hou2024universal, cho2024position, cho2024arithmetic} achieved non-trivial 2--3x length generalization with high but not perfect exact-match accuracy (EMA).

Interestingly, similar to us, \citet{hou2024universal} train on scratchpad data consisting of the computational traces of a Turing machine (TM) designed to solve the specific algorithmic task. However, whereas our agent only observes the evolution of the TM tape, \citet{hou2024universal} train with access to the TM state transitions, effectively trivializing the task, as the transformer merely needs to memorize the given algorithm without performing any program synthesis.

Finally, we make a detailed comparison in \Cref{tab:comparison} with the prior work \citep{kaiser2015neural, price2016extensions} most relevant to our paper. \citet{kaiser2015neural} introduce a Neural Cellular Automaton, called the Neural GPU, which achieves remarkable PLG when trained solely on input-output examples of binary addition and multiplication. Unfortunately, \citet{price2016extensions} report that the Neural GPU's generalization capacity highly depends on the random seed used for training and initialization. Only a few random seeds and hyperparameter configurations lead to 2000-digit PLG, and all trained models fail on highly structured and symmetric examples containing only a few digits. In contrast, our trained models achieve perfect accuracy on long symmetric examples as far as we could test, and unlike the Neural GPU, also manage to learn the challenging decimal arithmetic tasks. Furthermore, our DFST training relies only on a simple cosine learning-rate scheduler, while the Neural GPU training depends on multiple intricate techniques, including curriculum learning, gradient-noise injection, gradient scheduling, relaxation-pull, dropout, and gate cut-off. Hence, as shown in \Cref{tab:comparison}, the use of observable intermediate steps increases the robustness and simplicity of the training procedure, greatly diminishes the number of training samples, and enables the learning of decimal arithmetic.
\vspace{-2.5mm}
\section{Learning Framework}\label{sec:framework}
\vspace{-2.5mm}
Intelligent agents in the wild hardly ever learn novel algorithmic tasks from input-output observations. Instead, real-life learners observe computational processes and reconstruct the ground-truth function by making sense of the intermediate steps. We briefly restate the formalization of this idea due to \citet{papazov2025learning} before adapting the framework to our setting. We start with the underlying environment, which we model as a symbolic universe.

\begin{definition}[Symbolic Universe] \label{def:universe}
    Given an enumerable set $G$ and a finite set of symbols $\Sigma$ containing the empty symbol $\l$, a world-state $w : G \to \Sigma$ is a function with finitely many non-empty assignments. The symbolic universe $\U = \U(G, \Sigma)$ is the set of all such world-states.
\end{definition}

The set $G$ serves as the geometry of the universe. For our setup, we will only work on an unbounded 2-dimensional grid $G = \Z^2$ and observe the computation of \textit{grid agents}.

\begin{definition}[Grid Agent] \label{def:sca}
    A grid agent operating in a symbolic universe $\U(\Z^2, \Sigma)$ constitutes a triple $M = (Q, \U, \d)$, where $Q$ is a finite set of states and $\d : Q \times \Z^2 \times \U \to Q \times \Z^2 \times \U$ is a computable transition function with perception restrictions given below.
\end{definition}
The grid agent $M$ evolves and interacts with the environment according to the deterministic transition rules $\d(q, p, w) = (q', p', w')$, where $q \mapsto q' \in Q$ denotes the state update, $p \mapsto p' \in \Z^2$ -- the shift in perception, and $w \mapsto w'\in \U$ -- the change of the world-state. Importantly, \textbf{(i)} $\d(q, p, w)$ depends only on the current state $q$ and the observed cell $w(p)$; \textbf{(ii)} $\Vert p-p' \Vert_2 \leq 1$ -- i.e., the agent moves \textit{continuously} up (U), down (D), left (L), right (R), or stays (S); and \textbf{(iii)} $M$ can only edit the world-state at position $p$. For a complete interaction trace, we specify an initial state $q_0 \in Q$, an initial perceived cell $p_0 \in \Z^2$, and a final state $q_f \in Q$ as a halting condition. Note that by setting $\M = \{U, D, L, R, S\}$, we can essentially redefine the transition function $\d$ as a mapping from $Q \times \Sigma$ to $Q \times \Sigma \times \M$. Clearly, grid agents extend Turing machines to the plane, and hence, the family of grid agents $\G$ is computationally universal.

Grid agents also naturally model how human teachers navigate around a blackboard, with $q$ corresponding to the teacher's state of mind, $p$ -- the chalk's position, and $w(p)$ and $w'(p)$ -- the currently observed and written symbol, respectively. This analogy allows us to derive the policy-trajectory observations defined in \citep{papazov2025learning}. Indeed, at time $t \in \N$, the students have no access to the teacher's current neural state $q_t$ but can still observe how the teacher $M$ reacts with a symbol token $s_t = w_t(p_t) \in \Sigma$ and a motion token $m_t =$ ``$p_{t+1}-p_t$'' $\in \M$ to the current partial observation $x_t = w_t(p_t) \in \Sigma$. Thus, the students observe the trajectory $(x_t, a_t)_{t=0}^T$ of the history-dependent policy $\pi_M(x_0, \dots, x_t) = a_t = (s_t, m_t)$ enacted by the teacher -- hence the name, PTOs.

In our code base, we constructed 4 grid agents $M_\texttt{add2}$, $M_\texttt{add10}$, $M_\texttt{mult2}$, $M_\texttt{mult10}$ emulating the aforementioned 4 arithmetic tasks and operating on 4 problem alphabets $\Sigma_\texttt{add2} = \{0,1,\l,+\}$, $\Sigma_\texttt{add10} = \{0,\dots,9,\l,+\}$, $\Sigma_\texttt{mult2} = \{0,1,\l,\times\}$, $\Sigma_\texttt{mult10} = \{0,\dots,9,\l,\times\}$.
Each of these grid agents receives as input a string $z = a \circ b$ (where $\circ = +/\times$) horizontally written on the grid, while all other cells remain empty, and proceeds to compute, leaving only the final answer on the grid. 

\begin{figure}[h!]
    \centering
    \includegraphics[width=0.60\textwidth]{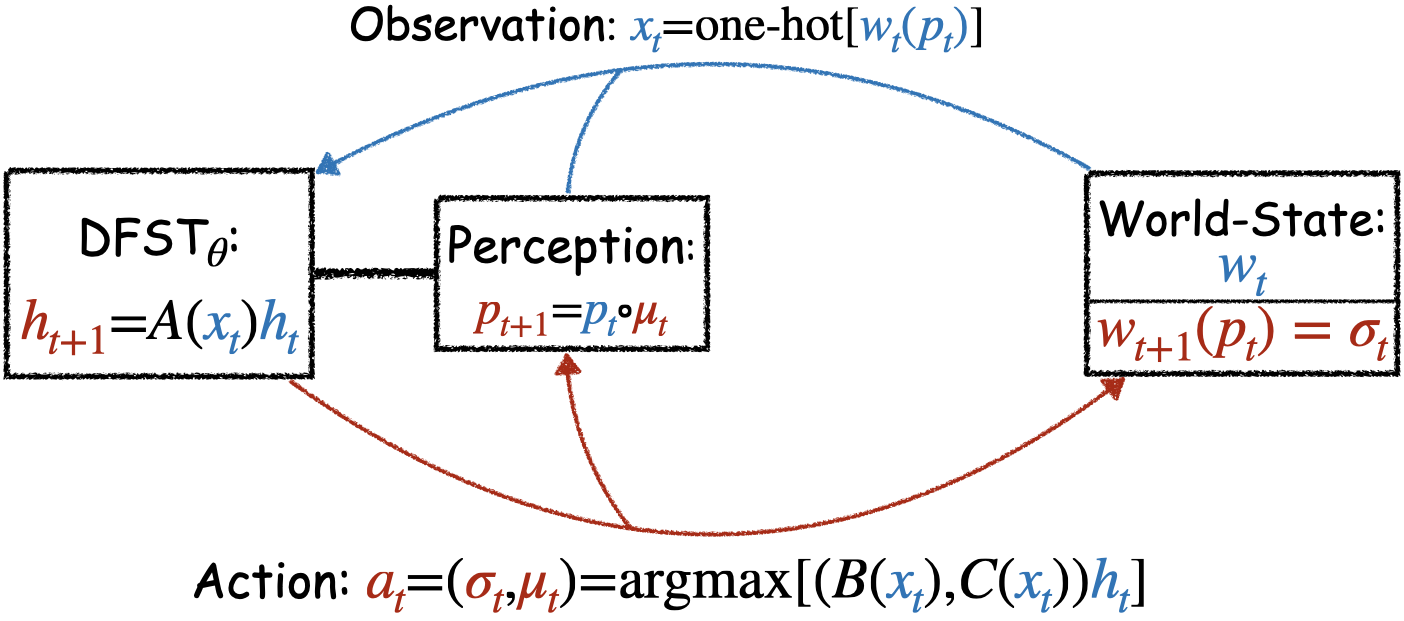}
    \caption{A DFST agent interacting with an external environment.}
    \label{fig:diff_agent}
\end{figure}

\textbf{A Differentiable Learner.} \citet{papazov2025learning} showed that a computationally costly learning-by-enumeration strategy provably identifies any computational agent $M^\star$ in the limit from PTOs. In this paper, we consider a more efficient gradient-based approach. Namely, we assume that the ground-truth model $M^\star$ comes from the universal class $\G$ of grid agents and train a DFST (defined below), on a small dataset of computational traces from $M^\star$.
\begin{definition}[Differentiable Finite-State Transducer Agent]
    Given a symbol-token alphabet $\Sigma$ and a motion-token alphabet $\M$, a DFST of dimension $d$ operating over the symbolic grid $\U(\Z^2, \Sigma)$ with motion vocabulary $\M$ consists of 3 trainable weight tensors $A \in \R^{|\Sigma| \times d \times d}$, $B \in \R^{|\Sigma| \times |\Sigma| \times d}$, $C \in \R^{|\Sigma| \times |\M| \times d}$ and a trainable initial hidden state $h_0 \in \R^d$. Upon observing a one-hot symbol encoding $x_t \in \R^{|\Sigma|}$ at time $t\geq 0$, which corresponds to the currently perceived part of the world-state $w_t(p_t)$, the DFST performs the following updates:
    \begin{align*}
        & h_{t+1} = A(x_t) h_t \\
        & \hat{s}_t = B(x_t) h_t \\
        & \hat{m}_t = C(x_t) h_t,
    \end{align*}
    and outputs a symbol token $\sigma_t = \argmax(\hat{s}_t) \in \Sigma$ and a motion token $\mu_t = \argmax(\hat{m}_t) \in \M$, which change the world-state and shift the perception as in \Cref{fig:diff_agent}. 
\end{definition}

\vspace{-2.5mm}

Let us denote by $\Delta_p$ the space of $p-$precision DFSTs. Let $\psi_p : \Delta_p \to \G$ denote the mapping that equates a DFST agent $M_\partial(A,B,C,h_0)$ with a grid agent $M(Q,\d)$ by building the set of states $Q$ from the $2^{pd}$ possible values for $h_t$ and by constructing the transition function $\d$ from the tensors $A, B, C$. Clearly, a DFST with a hidden-state dimension $d$ can express the policy of any grid agent with at most $d$ states by one-hot encoding the state transitions $\d$ into the tensors $A, B, C$. In other words, if $M \in \G$ has $d$ states $q_0, \dots, q_{d-1}$, $\Sigma$ has $k$ symbols $s_0, \dots, s_{k-1}$, and $\M$ has $r$ motion tokens $m_0, \dots, m_{r-1}$, we let $q_i, s_i,$ and $m_i$ correspond to the basis vectors $e_{i+1}$ in $\R^d, \R^k$, and $\R^r$, respectively. Then, we set $h_0^\d = e_1$, and if $\d(q_i, s_j) = (q_{i'}, s_{j'}, m_\ell)$, we set $A^\d[j, i', i] = 1, B^\d[j, j', i] = 1, C^\d[j, \ell, i] = 1$, and we define all other tensor weights as 0. Thus, the updates issued by the DFST($A^\d, B^\d, C^\d, h_0^\d$) exactly emulate $\d$, and we arrive at the following theorem.
\begin{theorem}[Universality of DFSTs] The map $\psi_p : \Delta_p \to \G$ is surjective. In particular, for any precision $p \in \N$, the DFST family $\Delta_p$ is Turing-complete when interacting with an external symbolic grid $\U(\Z^2, \Sigma)$. Moreover, any $d-$state grid agent admits emulation by DFST agents of dimension $d$.
    
\end{theorem}
The novelty in our proposed model family comes from the decoupling of the differentiable agent and the environment. Indeed, sequential models such as Transformers and RNNs move in a single direction with no means of controlling the next interaction spot with the environment. We amend this limitation by allowing our DFST control unit to not only edit the grid but also to issue commands for where the next interaction should occur.

We emphasize the point that any sequential neural network capable of state tracking (i.e., emulating a finite-state transducer) could, in principle, learn the policy of a grid agen $M^\star \in \G$. Such models include RNNs, LSTMs, GRUs \citep{alon1991efficient, siegelmann1996recurrent, weiss2018practical, merrill2019sequential, svete2023recurrent} but not Transformers \citep{merrill2023parallelism} or diagonal and non-gated SSMs \citep{merrill2024illusion}. Nevertheless, we chose the DFST as our trainable model due to its linear structure and simplicity, which we hope will inspire future theoretical analysis. Moreover, since DFSTs contain no nonlinearities, parallel scans (à la \citet{blelloch1990prefix}) allow for \textbf{fast log-parallel training}.

\vspace{-2.5mm}
\section{Experimental Details}\label{sec:exp}
\vspace{-2.5mm}
We train on PTOs from the expert grid agents $M_{\texttt{task}R}$ with the MSE loss 
$\frac{1}{2T}\sum_{t=0}^T [(\hat{s}_t-s_t)^2+(\hat{m}_t-m_t)^2]$
under the next-action prediction (NAP) objective. Here, $s_t$ and $m_t$ denote the one-hot encodings of the symbol and motion tokens emitted by the expert models at time $t \geq 0$.

\textbf{Identity Initialization.}
We train four DFST control units $D_\texttt{add2}, D_\texttt{add10}, D_\texttt{mult2}$, and $D_\texttt{mult10}$ with hidden-state dimensions matching the number of states of the corresponding ground-truth grid agent: i.e., 12, 20, 32, and 108. 
We initialize the state-transition matrices $A[i], \forall i < |\Sigma|,$ as the identity $I_d \in \R^{d \times d}$ and the tensors $B$ and $C$ as 0. We initialize each coordinate of $h_0$ uniformly at random and independently in the interval (0,1), after which we normalize $h_0$ to have a unit Euclidean norm.

\textbf{Optimizer.}
We use the standard Adam optimizer with a cosine annealing scheduler and no warm-up period. We start the training of $D_\texttt{add10}, D_\texttt{mult2}$, and $D_\texttt{mult10}$ with a learning rate of 0.001, and the training of $D_\texttt{add2}$ -- with a learning rate of 0.01. We always use batch size 32 and \texttt{float32} precision.

\textbf{Data Selection.} We create a data sampling function $\texttt{DS}(R,\texttt{task}, p,q,N)$ which for radix $R$ and arithmetic task \texttt{task}, samples all PTOs from $M_{\texttt{task}R}$ on number pairs $(a,b)$ with at most $p$ digits -- that is, a total of $R^2(R^p-1)^2/(R-1)^2$ samples. Then, \texttt{DS} samples all $R^2$ pairs of $q$-digit numbers made of a single digit repeated $q$ times. After that, if $N > R^2(R^p-1)^2/(R-1)^2 + R^2$, \texttt{DS} samples at random another $N - R^2(R^p-1)^2/(R-1)^2 + R^2$ unique pairs of numbers having up to $q$ digits. For $\texttt{add2}$ we use $p=1,q=3,N=20$, for $\texttt{add10}$ --- $p=1,q=3,N=225$, for $\texttt{mult2}$ --- $p=1,q=5,N=750$, and for $\texttt{mult10}$ --- $p=1,q=5,N=10000$.

\textbf{Length Generalization.} We define the Probabilistic Length Generalization (PLG) of a model as the largest number of digits $m$ such that the model achieves perfect EMA when tested on 5 random pairs of numbers having exactly $m$ digits and 5 random pairs of numbers having at most $m$ digits. We define the Robust Length Generalization (RLG) of a model as the largest number of digits $m$ such that $\text{PLG} \geq m$ and the model achieves perfect EMA on the $R^2$ pairs of $m$-digit numbers that have $m$ identical digits. We noticed after extensive testing that (as observed by \citet{price2016extensions}) same-digit numbers constitute the hard test instances for generalization. Clearly, $\text{RLG} \leq \text{PLG}$. In \Cref{app:plots}, we show the relationship between longer training time and RLG for experiments with random seed 42. In \Cref{tab:comparison}, our reported values for PLG and RLG reflect our experiments on \texttt{add2}, \texttt{add10}, \texttt{mult2}, and \texttt{mult10} for training lengths 500k, 500k, 3mln, and 3mln iterations, respectively.

\textbf{Memory Constraints.} During training, our DFST models observe PTOs of lengths at most 70 for the addition tasks and at most 464 for the multiplication tasks. However, the lengths of PTOs from $M_\texttt{add2}$, $M_\texttt{add10}$, $M_\texttt{mult2}$, $M_\texttt{mult10}$ on number pairs with 3850, 2450, 600, and 180 digits, respectively, become on the order of 30mln, 12mln, 6mln, and 500k operations. Checking the EMA on sequences above these lengths completely fills the memory of our A100-SXM4-80GB GPU.

\textbf{Verifying Exact Learning.} We draw attention to the fact that verifying whether two grid agents follow the same deterministic policy reduces to deciding the equivalence of two Turing machines. Since the language $\mathrm{EQ}_{\mathrm{TM}} = \big\{\, \langle M_1, M_2 \rangle \;\big|\; L(M_1) = L(M_2) \,\big\}$ is undecidable, no general procedure can check whether $\pi_{M_{\texttt{task}R}} \equiv \pi_{D_{\texttt{task}R}}$ by just observing the descriptions of $M_{\texttt{task}R}$ and $D_{\texttt{task}R}$. Consequently, we rely on random tests to gain confidence in the ability of $D_{\texttt{task}R}$.
\vspace{-2.5mm}
\section{Conclusion}\label{sec:conclusion}
\vspace{-2.5mm}
Our results demonstrate that Differentiable Finite-State Transducers achieve unprecedented length generalization on arithmetic tasks using simple, gradient-based training. This work provides a concrete step toward the broader goal of enabling differentiable agents to learn precise algorithmic behavior through structured intermediate supervision. We hope that our minimalist, differentiable, and Turing-complete framework will inspire further theoretical explorations of the loss landscape of algorithmic learning.

\bibliography{references}


\appendix
\section{Robust Length Generalization vs. Training Time}\label{app:plots}
In this section we report the experimental results that highlight the trade-off between robust length generalization and training time across various model sizes and training tasks. Our findings indicate that increased training times correlate with lower training loss and improved length generalization capabilities.
\begin{figure}[h]
    \centering
    \includegraphics[width=1.0\linewidth]{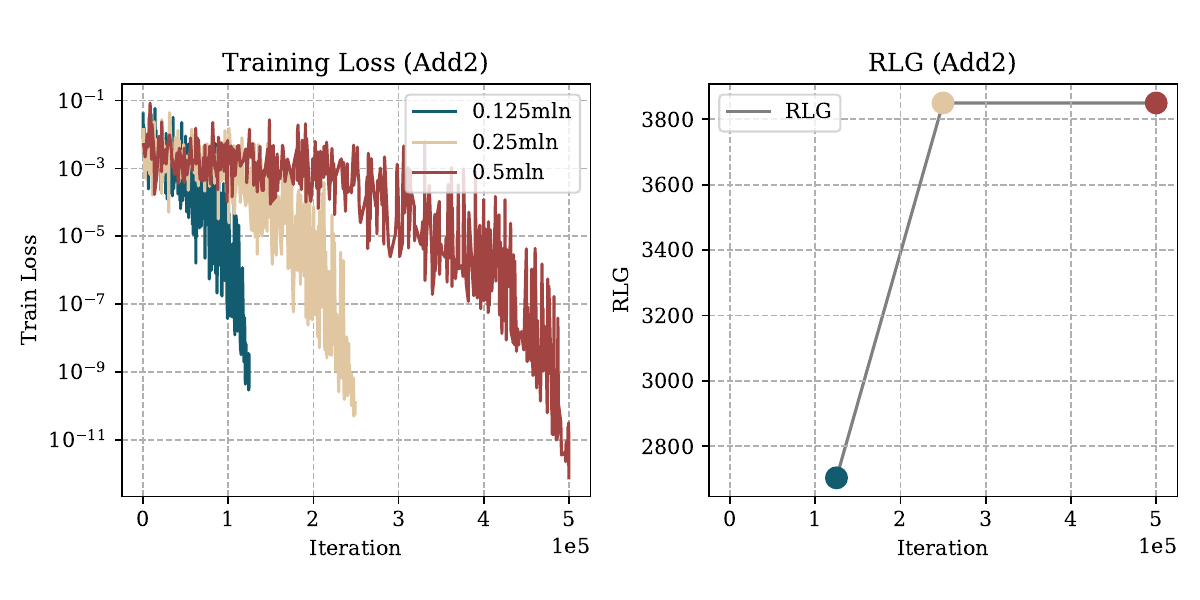}
    \caption{\textbf{Binary addition:} Training loss (left) and RLG (robust length generalization, right) across training iterations.}
    \label{fig:app_add2}
\end{figure}
\begin{figure}[h]
    \centering
    \includegraphics[width=1.0\linewidth]{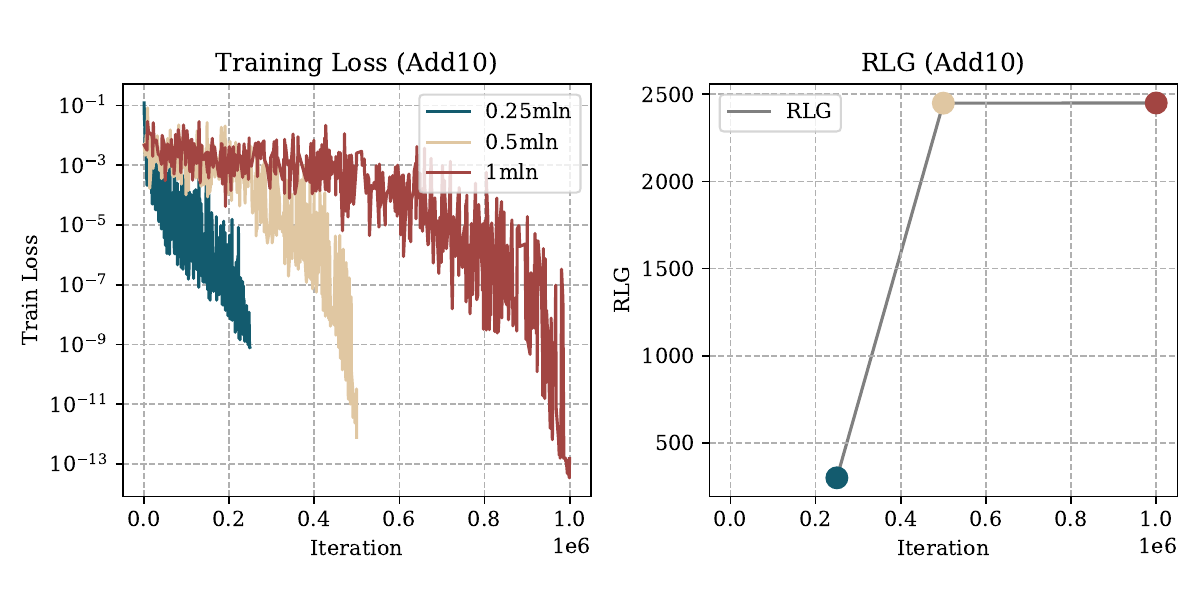}
    \caption{\textbf{Decimal addition:}  Training loss (left) and RLG (robust length generalization, right) across training iterations.}
    \label{fig:app_add10}
\end{figure}
\begin{figure}[h]
    \centering
    \includegraphics[width=1.0\linewidth]{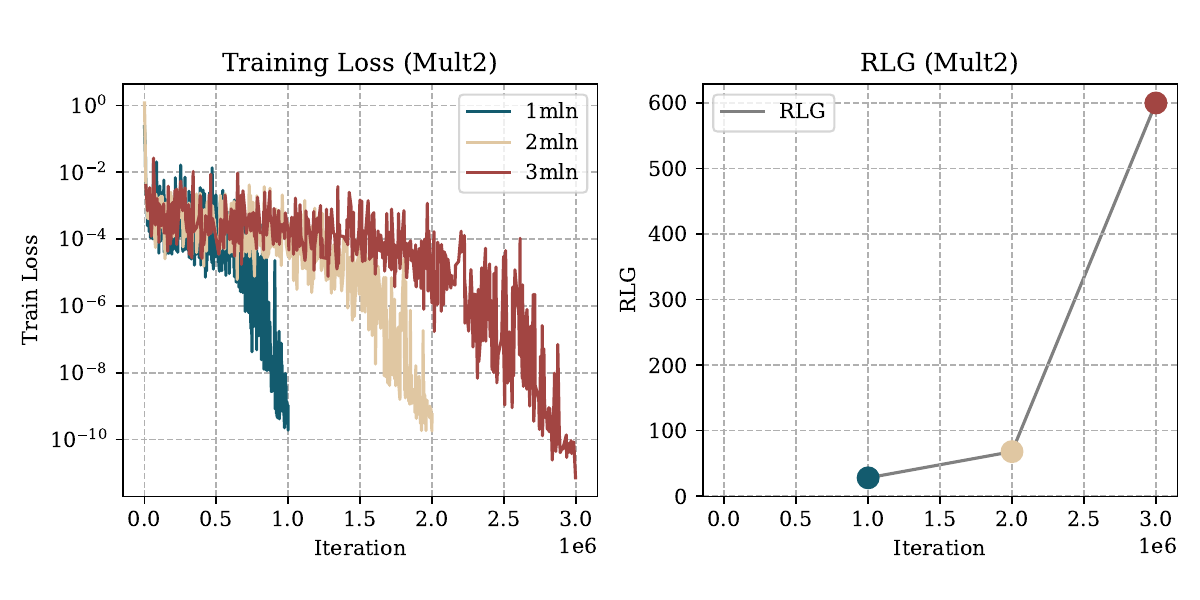}
    \caption{\textbf{Binary Multiplication:}  Training loss (left) and RLG (robust length generalization, right) across training iterations.}
    \label{fig:app_mult2}
\end{figure}
\begin{figure}[h]
    \centering
    \includegraphics[width=1.0\linewidth]{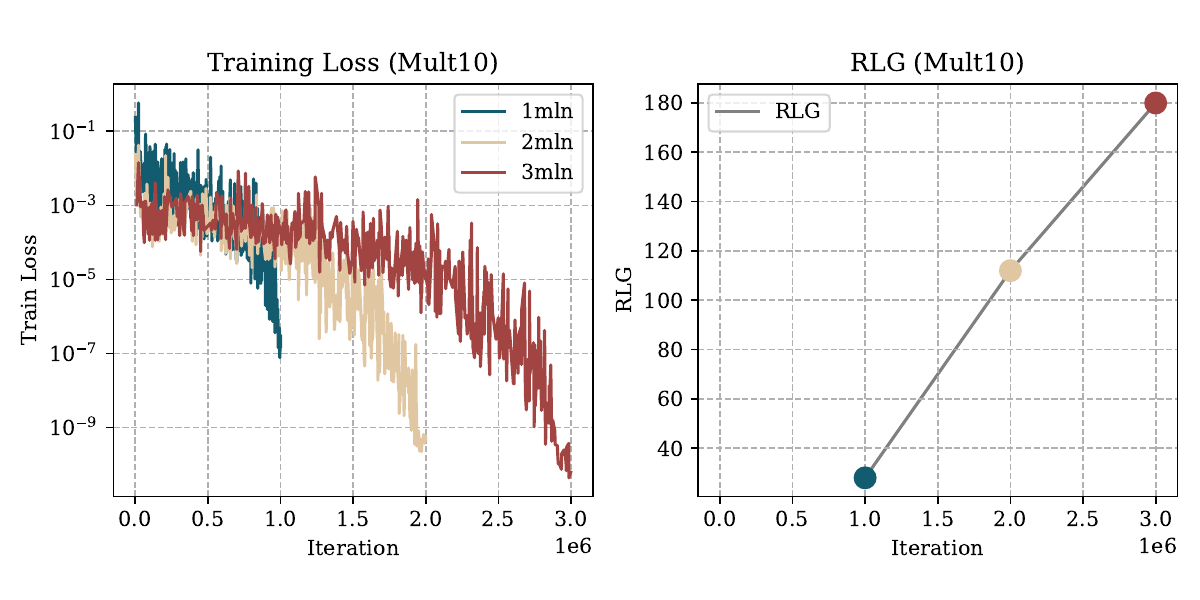}
    \caption{\textbf{Decimal Multiplication:}  Training loss (left) and RLG (robust length generalization, right) across training iterations.}
    \label{fig:app_mult10}
\end{figure}


\end{document}